\documentclass[10pt,letterpaper]{article}

\usepackage{cogsci}
\usepackage{caption}
\cogscifinalcopy
\usepackage{pslatex}
\usepackage{apacite}
\usepackage{svg}
\usepackage{graphicx}
\usepackage{multicol}
\usepackage{float} 
\title{Long-form analogies generated by chatGPT lack human-like psycholinguistic properties}
 
\author{{\large \bf S.M. Seals (seals.25@wright.edu)\textsuperscript{1}} \\ \textsuperscript{1} Wright State University Department of Psychology \\ 335 Fawcett Hall, 3640 Colonel Glen Hwy \\  Dayton, OH 45435 USA 
     \And
        {\large \bf Valerie L. Shalin (valerie.shalin@wright.edu)\textsuperscript{1,2}}\\ \textsuperscript{2} AI Institute University of South Carolina \\ Columbia, SC 29208 USA}

\begin{document}

\maketitle

\begin{abstract}
    Psycholinguistic analyses provide a means of evaluating large language model (LLM) output and making systematic comparisons to human-generated text. These methods can be used to characterize the psycholinguistic properties of LLM output and illustrate areas where LLMs fall short in comparison to human-generated text. In this work, we apply psycholinguistic methods to evaluate individual sentences from long-form analogies about biochemical concepts. We compare analogies generated by human subjects enrolled in introductory biochemistry courses to analogies generated by chatGPT. We perform a supervised classification analysis using 78 features extracted from Coh-metrix that analyze text cohesion, language, and readability \cite{graesser_coh-metrix_2004}. Results illustrate high performance for classifying student-generated and chatGPT-generated analogies. To evaluate which features contribute most to model performance, we use a hierarchical clustering approach. Results from this analysis illustrate several linguistic differences between the two sources.

\textbf{Keywords: Large Language Models; chatGPT; analogical reasoning; psycholinguistics} 
\end{abstract}

\section{Introduction}
Natural language processing and deep learning have enabled the development of personal assistants \cite{radford_language_2019}, automatic speech recognition, \cite{hinton_deep_2012}, machine translation systems \cite{sutskever_sequence_2014}, and large language models that can generate convincing human-like text and dialogue \cite{brown_language_2020, devlin_bert_2019, ouyangTrainingLanguageModels2022, zhangOPTOpenPretrained2022}. One such model that has attracted considerable scientific and popular attention is chatGPT, a large language model released by openAI in December of 2022 \footnote{https://openai.com/blog/chatgpt/}.

In this paper we examine the potential differences between ChatGPT output and human output.  We examine performance on a task regarded as indicative of sophisticated human reasoning:  the generation of analogies in support of scientific reasoning.  We employ computationally grounded metrics established in psycholinguistic research \cite{graesser_coh-metrix_2004}.

\section{Background}

Language and reasoning are commonly considered to require human-like cognitive capabilities. Recent advances in computational natural language processing motivate careful analysis of the potential differences between the output of such processes and human behavior using features that are sufficiently sensitive to subtle differences in language abilities. 

Language models represent a probability distribution of sequences of words, content and syntax in large natural language corpora \cite{jurafsky_speech_2008}.  Models are trained on input text, typically represented as words, morphemes, or characters and aim to capture underlying statistical properties of language in the input text. Language models do not generate text directly, rather they output probabilities that can be used to select words. Recent advances in natural language processing and deep learning have enabled the development of language models that are considerably larger than their  predecessors and can generate text that can, in some instances, pass for that of a human author \cite{brown_language_2020, devlin_bert_2019, shoeybiMegatronLMTrainingMultiBillion2020}. Here we ask the extent to which LLMs are plausible models of skilled human language. 

\subsection{What is chatGPT?}
InstructGPT is a large language model (LLM) recently released by openAI that was designed to follow 
natural language based instructions better than its predecessor GPT-3 \cite{ouyangTrainingLanguageModels2022}. One of the known limitations of LLMs is the generation of undesirable content. For example, LLMs may "make up" facts, generate biased or toxic text, or simply not comply with user instructions \cite{Bender2021, gehman_realtoxicityprompts_2020, ji_survey_hallucination_2022, lu_gender_2019}. InstructGPT uses human feedback to minimize such undesirable content.

ChatGPT is a 'sibling' model of instructGPT that is currently available to use during a research preview \footnote{https://chat.openai.com/chat}. ChatGPT was trained using a reinforcement learning paradigm that incorporates human feedback \footnote{https://openai.com/blog/chatgpt/}. The initial model was trained with supervised fine-tuning. Humans provided sample conversations to the model where both the example AI assistant and the human participant dialogue were written by humans. Human trainers were provided with suggested model-generated responses to assist with composing responses. This dataset was combined with the dataset for InstructGPT (which was converted to dialogue format). The reward function for chatGPT was created by randomly sampling alternative completions for a given prompt and having humans rate the quality of the alternative completions. This information was used to fine-tune the model with proximal policy optimization \cite{schulmanProximalPolicyOptimization2017}. This version is the source of LLM text for the present analysis.

\subsection{Analysis Method Background}
We employed a machine learning classifier to distinguish between ChatGPT text and human-generated text. Supervised methods employ labeled data in a training process. Unsupervised methods do not require such labels and simply attempt to group like material with like. For some AI-detection problems, ground truth labels may not be available \cite{mitchellDetectGPTZeroShotMachineGenerated2023}. However, due to the design of this experiment, ground truth labels are available and we use a supervised method to characterize the differences between the two sources of data.

\subsection{Psycholinguistic Analysis of Text}
While many ML efforts depend on vector representations of material to be clustered, the resulting findings are difficult to interpret or provide theoretical insight.  Therefore, we employ psychologically motivated "engineered" features to describe text, which if diagnostic, explain the differences between clusters.  
We use Coh-Metrix, an automated tool that calculates computational metrics of cohesion and coherence for written and spoken texts \cite{graesser_coh-metrix_2004}. Coh-Metrix includes measures that capture word information and frequency, syntactic complexity, polysemy and hypernymy, the frequency of major parts of speech, use of connectives and logical operators, and cohesion \cite{graesser_coh-metrix_2004, McCarthy2006}. A large research base in psycholinguistics, cognitive psychology, and education informs the identification of relevant features \cite{ Coltheart1981, graesser_coh-metrix_2004, halliday2014cohesion, just_theory_1980, mcnamara_importance_2012, miller1998wordnet, VanDijk1983}. 

Coh-Metrix is particularly relevant here for its original focus on instructional applications.  Coh-Metrix  has been used in other applications as well. For example, previous work demonstrated the use of Coh-Metrix for classifying adults as healthy elderly controls, patients with Mild Cognitive Impairment, patients with possible Alzheimer's disease, and patients with probably Alzheimer's disease \cite{padhee_predicting_2020}. 

\section{Methods}
In the following sub-sections, we discuss the data, psycholinguistic features, and the analysis method we used. 

\subsection{Data}

We used two sets of data, a human subject sample and a chatGPT sample. Participants in the original study were enrolled in three different biochemistry courses taught at a large southeastern university in the United States. The original data collection was reviewed and approved by the Institutional Review Board. Subsequent analysis occurred on de-identified data. The study examined two different biochemical processes, glycolysis and enzyme kinetics. Instructors first introduced the relevant topics in class by assigning textbook readings and conducting lectures on the relevant biochemical process. Students were then provided with a sample analogy to explain the relevant biochemical process. Participants were then asked to create their analogy to explain the biochemical process using a subject of their choice. This unique task is the foundation of EngageFastLearning \footnote{https://www.engagefastlearning.com/}. Participants created a total of 500 analogies in this study. To control for class balance and power for this project, we sampled ($n = 31$) from the 500 for analysis. 

For each biochemical process, we prompted chatGPT to generate analogies. For the enzyme kinetics example, we gave chatGPT the prompt: \emph{Create an analogy to explain how enzyme kinetics works.} To generate the chatGPT glycolysis examples, we gave chatGPT the prompt: \emph{Create an analogy to explain how glycolysis works.} We found that chatGPT will create multiple versions of the same response if an identical prompt is repeated. On subsequent requests for each topic, we replaced \emph{Create an analogy} with \emph{Create a new analogy}. We generated a total of 51 chatGPT analogies, all of which were created on December 20th, 2022.

\subsection{Psycholinguistic Features}
We calculated features using Coh-metrix version 3.0. Coh-metrix generates a total of 108 features. We removed a total of 30 Coh-Metrix features with low variance ($\sigma^2 < 0.01$), retaining 78. We scaled the features to have a mean of zero and standard deviation of 1.

\subsection{Analysis Method}

We use a linear ridge classifier to analyze our data. Ridge is a supervised learning method that imposes a penalty on the size of the regression coefficients \cite{Hastie2009}. This penalty is intended to reduce overfitting and increase performance on unseen data. 

\begin{figure}[h]
    \centering
    \includegraphics[width = 0.48\textwidth]{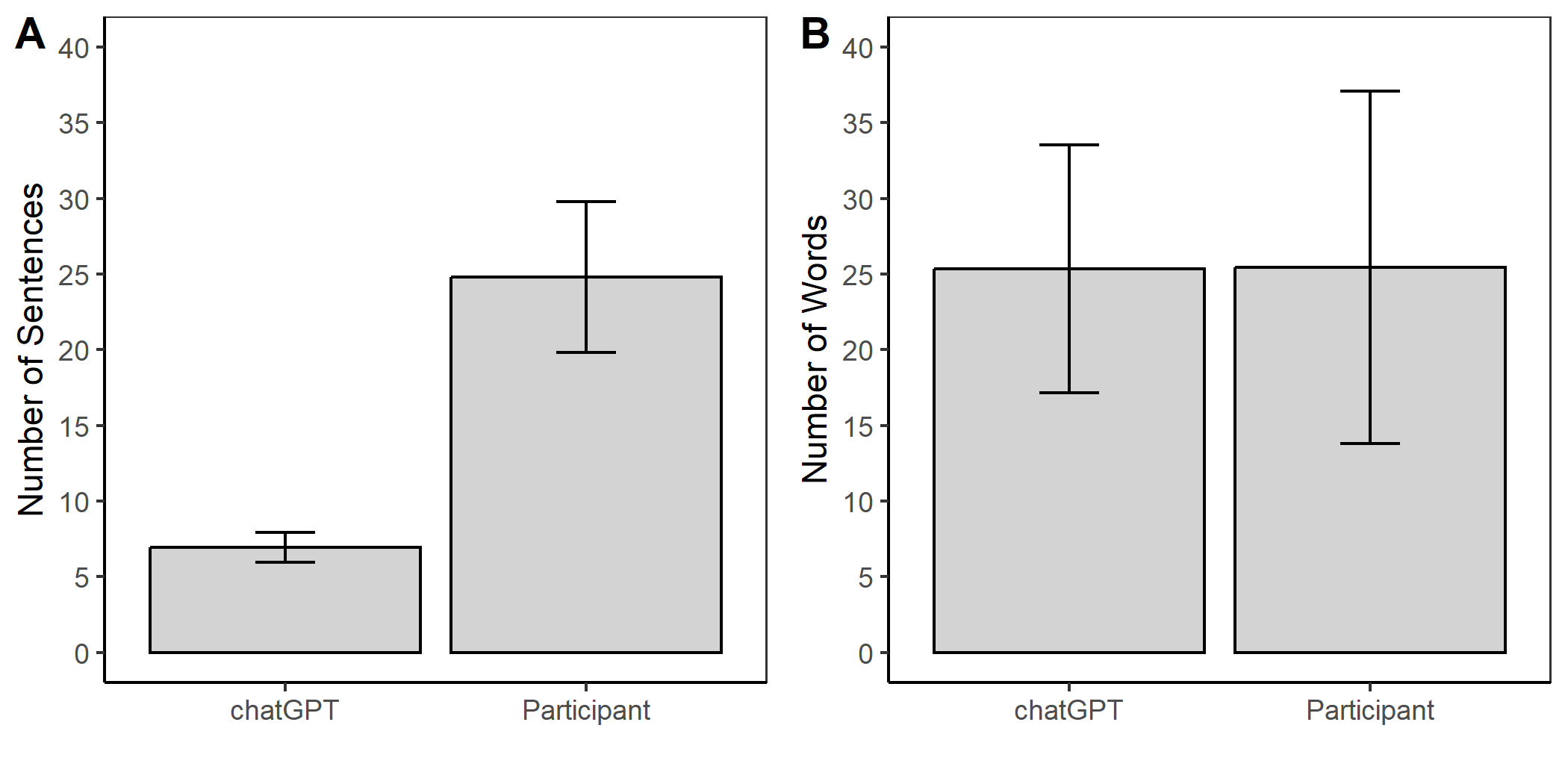}
    \caption{A: Number of sentences by analogy source. chatGPT: M = 6.92, SD = 0.98; Subjects: M = 24.8, SD = 4.96 B: Number of words by analogy source. chatGPT: M = 25.3, SD = 8.17; Subjects: participants: M = 25.4, SD= 11.6. Error bars in both represent +/- 1 SD.}
    \label{fig:desc_stats}
\end{figure}

Initial inspection of the data indicated that the chatGPT analogies differ in length from those generated by subjects (see Figure \ref{fig:desc_stats}). Notably, the mean length of an analogy generated by chatGPT is shorter than those generated by the subjects ($t = 17.35, df = 33, p < 0.001$). Fitting a classification algorithm on these two samples would allow the algorithm to capitalize on analogy length in order to classify the examples. Thus, we perform our analysis on individual sentences to control for this. Moreover, this is consistent with prior work examining NLP analysis of analogies \cite{wijesiriwardene_towards_2022}. This results in a total of 366 chatGPT sentences and 735 participant sentences. 

We report three sets of analyses. In the first set of analyses, we evaluate the performance of a model fit on all 78 features. 

In the second set of analyses, we remove two linguistic features: the length of each sentence (in words) and the standard deviation of sentence length. As illustrated in Figure~\ref{fig:desc_stats} B, the mean sentence length is similar for the subjects and chatGPT. However, the standard deviation of sentence length is smaller for chatGPT than for the participants ($F$ variance test: $F = 0.47, df_n = 365, df_d = 734, p < 0.001$).

In the third set of analyses, we remove all features associated with length. These include total word count, sentence length, sentence length standard deviation, word length, and word length standard deviation. 

For each set of analyses, we fit the model via a cross-validation procedure. We divide the data into ten stratified groups such that each group will contain roughly similar instances of both classes. We train the model on 9 of the groups, validate the model on the 10th group and repeat this procedure ten times. We report means and standard deviations for each metric.

\subsection{Feature Analysis Method} \label{sec: feature_analysis_method}

\begin{figure}[h]
    \centering
    \includegraphics[width = 0.75\columnwidth]{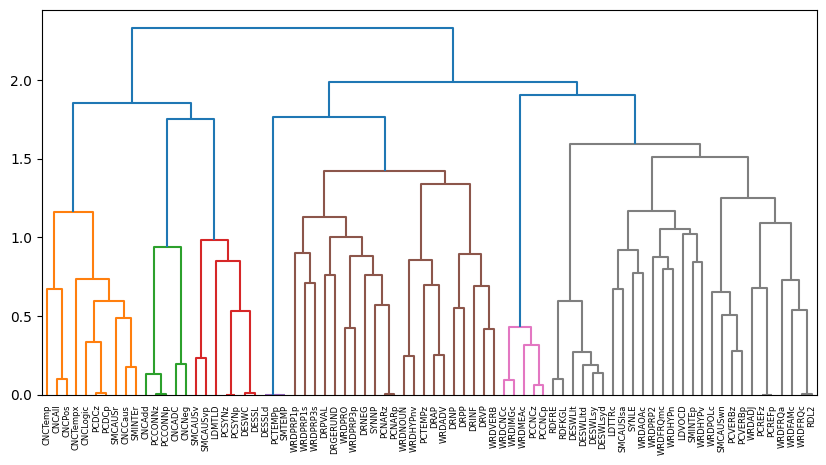}
    \caption{Hierarchical clustering dendrogram}
    \label{fig:dendro}
\end{figure}

For each set of analyses, we perform a feature analysis procedure to examine which features are most informative for classifying the data. Because many of the features available in Coh-Metrix are highly correlated, we use a hierarchical clustering method to examine the relative importance of the Coh-Metrix features for each model.

Hierarchical clustering algorithms use successive merging or splitting techniques to create nested clusters that can be represented in tree-like structures or dendrograms \cite{nielsen2016hierarchical}. First, we compute a Spearman rank correlation matrix for the features. We then convert this to a distance matrix. Next, we perform hierarchical clustering using a bottom up approach where each feature begins as a unique item and features are successively merged. We use the Ward criterion, which minimizes the sum of squared errors between the clusters \cite{ward1963hierarchical} as implemented here \footnote{https://scipy.org/}. A an illustration of the dendrogram for Analysis 1 appears in Figure ~\ref{fig:dendro}.

Next, we create flat clusters using the Wald criterion. We specify that the clusters cannot have a distance greater than 1.25 or half the cophenetic distance of the dendrogram illustrated in Figure ~\ref{fig:dendro}. Conceptually speaking, we take the features represented in the top half of the dendrogram.

Next, we use the selected features to perform another $k$-means cluster analysis with 2 clusters and random initial centroids. We perform a 10-fold cross validation procedure to generalize the results.

\section{Results}
Below, we discuss results for each of our three sets of analyses.

\subsection{Analysis 1}

In the first set of analyses, we evaluate the performance of a model that uses all 78 features. Results from this model are illustrated in the \textit{top half} of Table~\ref{tab:analysis1}. Results from this model indicate high performance at classifying the data as either created by subjects or created by chatGPT.

\begin{table}[h]
\begin{center} 
\caption{Analysis 1 Results} 
\label{tab:analysis1} 
\vskip 0.12in
\begin{tabular}{lll} 
\hline
Model & Metric  &  Score: M (SD) \\
\hline
Classification & Balanced Accuracy & 0.89 (0.08) \\
Classification & Weighted Precision & 0.90 (0.05) \\
Classification & Weighted Recall & 0.89 (0.05) \\
\hline
Feature & Balanced Accuracy & 0.70 (0.05) \\
Feature & Weighted Precision & 0.75 (0.04) \\
Feature & Weighted Recall & 0.75 (0.04) \\
\hline
\end{tabular} 
\caption*{\emph{Note.} \textbf{Top:} Results from ridge classifier fit on 78 Coh-metrix features. \textbf{Bottom:} Results from ridge classifier fit on 11 Coh-metrix features. All result metrics are means and standard deviations from 10-fold cross validation. 
Precision: ratio of true positives to sum of true positives and false positives, weighted by the frequency of each class. Recall: ratio of true positives to sum of true positives and false negatives, weighted by the frequency of each class. Balanced accuracy: average recall for each class.}
\end{center} 
\end{table}

Next, we use the hierarchical clustering approach described above to evaluate which features are most informative for classifying the data. We identified 11 features using this procedure. 

\begin{figure}[h]
    \includegraphics[width = \columnwidth]{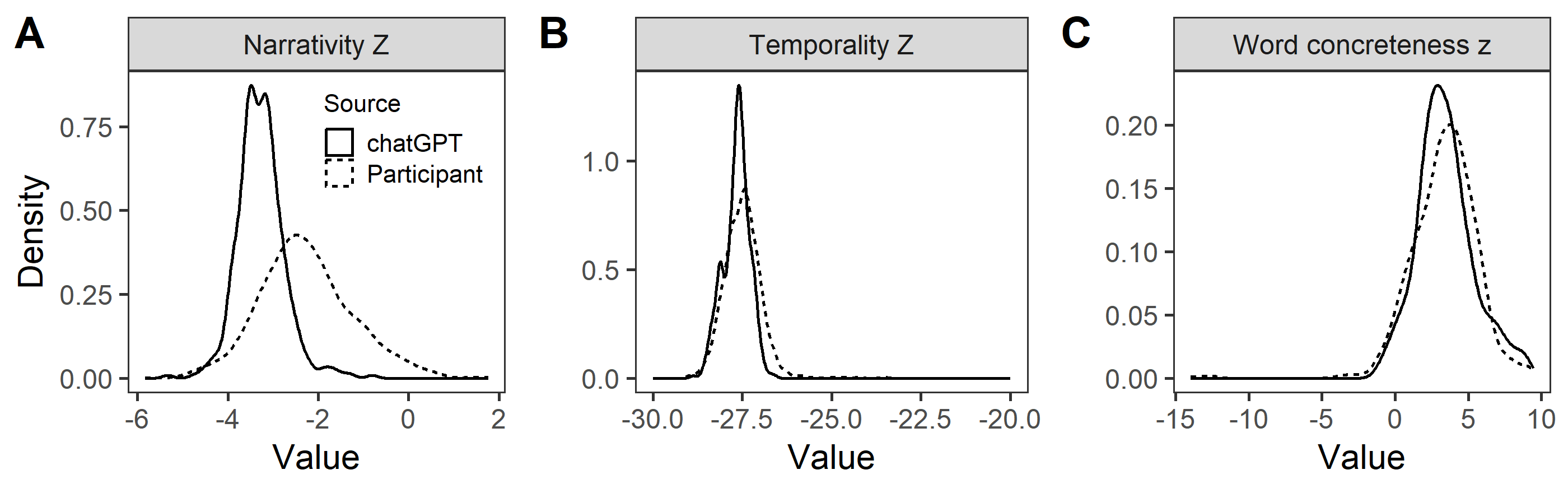}
    
    \includegraphics[width = \columnwidth]{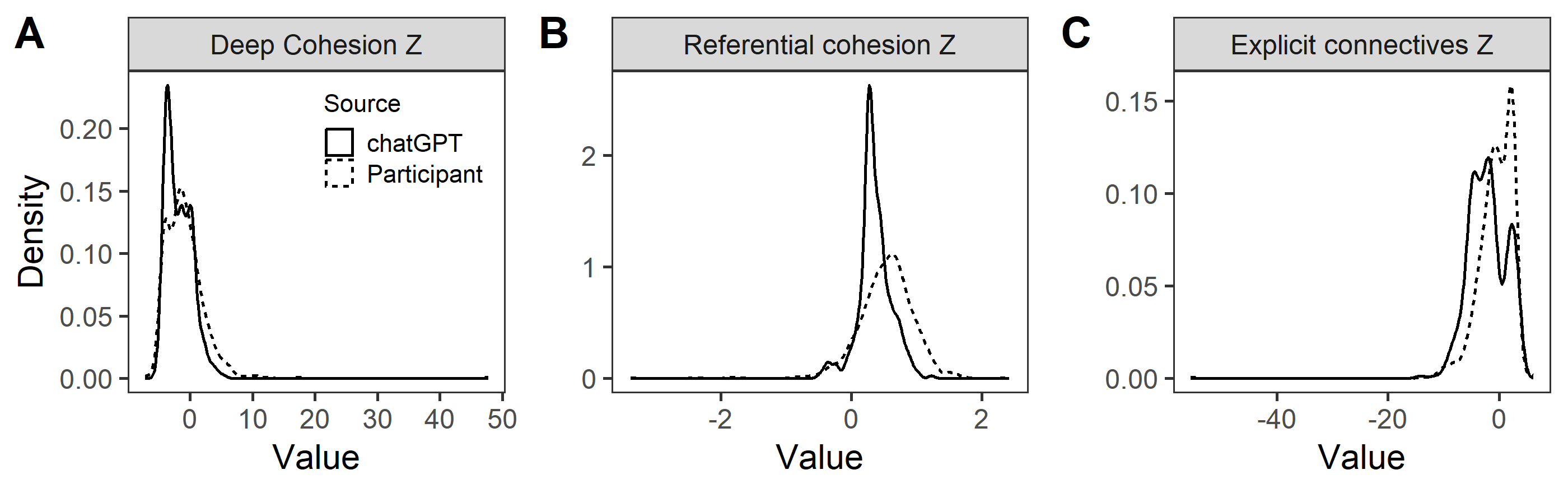}

    \includegraphics[width = 0.66\columnwidth]{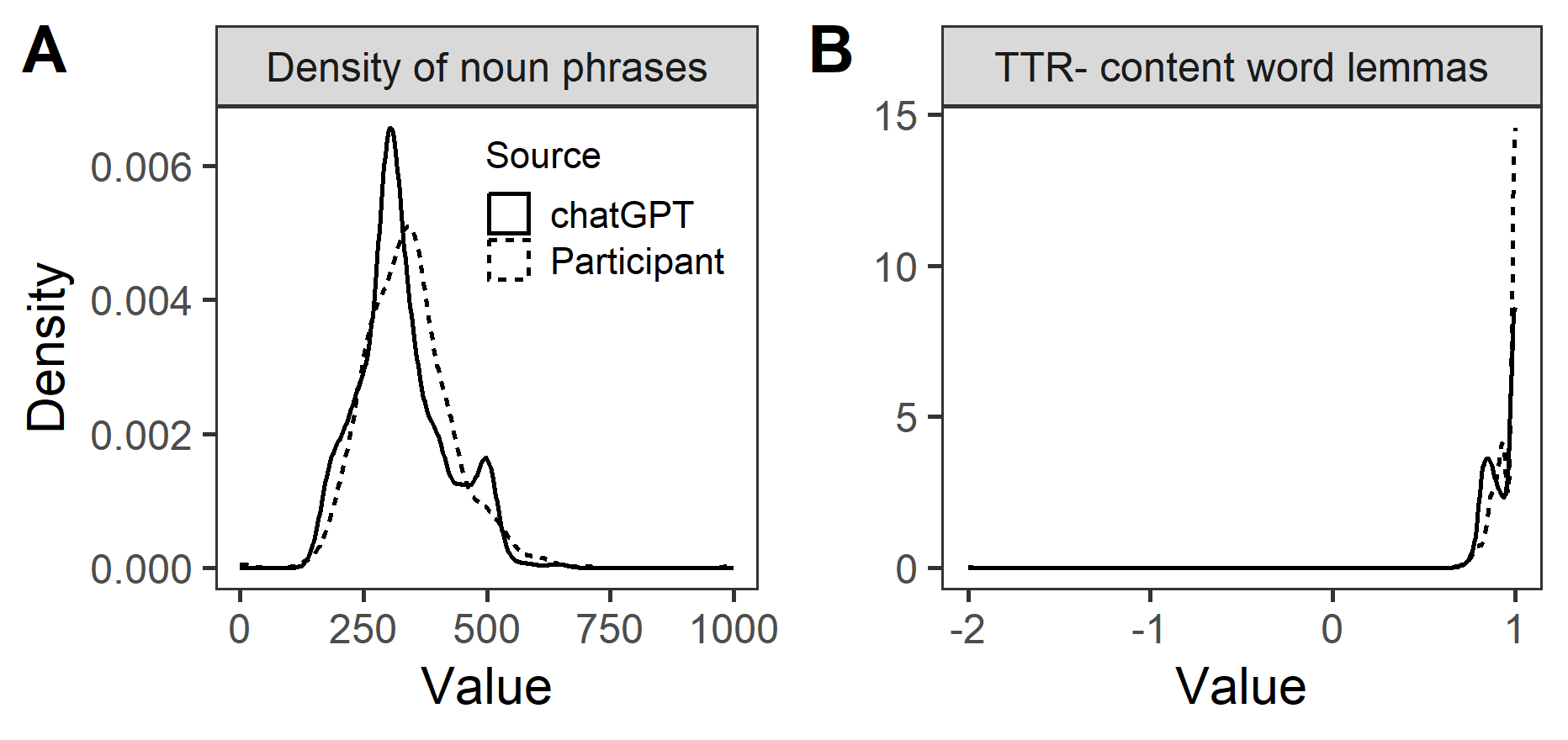}
    \caption{Distribution plots for: \textbf{Top row:} noun phrase density (\textbf{A}) and type token ratio for content word lemmas (\textbf{B}). \textbf{Middle row:} Z scores of narrativity (\textbf{A}), temporality (\textbf{B}), and concreteness (\textbf{C}). \textbf{Bottom row:} z scores of deep cohesion (\textbf{A}), referential cohesion (\textbf{B}), and the presence of explicit connectives (\textbf{C}).}
    \label{fig:a1_dists}
\end{figure}

These features are: total word count, sentence length, sentence length standard deviation, the type token ratio of content word lemmas, noun phrase density, and six features associated with text easability. The text easability features are designed to capture text difficulty in a more comprehensive manner than traditional grade level reading metrics (i.e., \citeNP{flesch_new_1948}). These measures provide readability information at different levels of discourse processing \cite{graesser_coh-metrix_2011}. The six easability features are z-scores for narrativity, word concreteness, referential cohesion, deep cohesion, connectivity, and temporality. Distribution plots for these variables appear in Figure \ref{fig:a1_dists}. 

Results from fitting a ridge classifier with these features are presented in the \textit{bottom half} of Table ~\ref{tab:analysis1}. Performance on this model is reduced compared to the full model (dependent $t$- test: Accuracy: $t = 7.64$, Precision: $t = 8.18$, Recall: $t = 7.86$, $df = 9$, $p < 0.001$). However, this model still illustrates reasonably high levels of classification performance with a much smaller feature set. 

\subsection{Analysis 2}
Analysis 1 illustrated that features associated with word count and sentence length were important to model classification performance. While there are clear differences in these features for our data set, these differences may not hold for new datasets. Thus, for analysis 2, we examine classification performance on a model that excludes these features. 

\begin{table}
\begin{center} 
\caption{Analysis 2 Results} 
\label{tab:analysis2} 
\vskip 0.12in
\begin{tabular}{lll} 
\hline
Model & Metric  &  Score: Mean (sd) \\
\hline
Classification & Balanced Accuracy & 0.88 (0.08) \\
Classification & Weighted Precision & 0.90 (0.06) \\
Classification & Weighted Recall & 0.89 (0.06) \\
\hline
Feature & Weighted Recall & 0.72 (0.06) \\
Feature & Weighted Recall & 0.76 (0.05) \\
Feature & Weighted Recall & 0.77 (0.04) \\
\hline
\end{tabular} 
\caption{\emph{Note.} \textbf{Top:} Results from ridge classifier fit on 75 Coh-metrix features (total word count, sentence length, and sentence length standard deviation are removed). \textbf{Bottom:} Results from ridge classifier fit on 11 Coh-metrix features selected via a hierarchical clustering approach based on Analysis 2. All result metrics are means and standard deviations from 10-fold cross validation.}
\end{center} 
\end{table}

We remove three linguistic features: total word count, the length of each sentence and the standard deviation of sentence length. Results from this model are illustrated in the top half of Table~\ref{tab:analysis2}. Results from this model are comparable to those reported in the top half of Table ~\ref{tab:analysis1} (dependent $t$- tests: Accuracy $t = 0.73$, Precision $t = 0.73$, Recall $t = 0.75$, $df = 9$, $p > 0.05$). This model demonstrates high performance at classifying the data as either created by subjects or created by chatGPT.

\begin{figure}
    \includegraphics[width = \columnwidth]{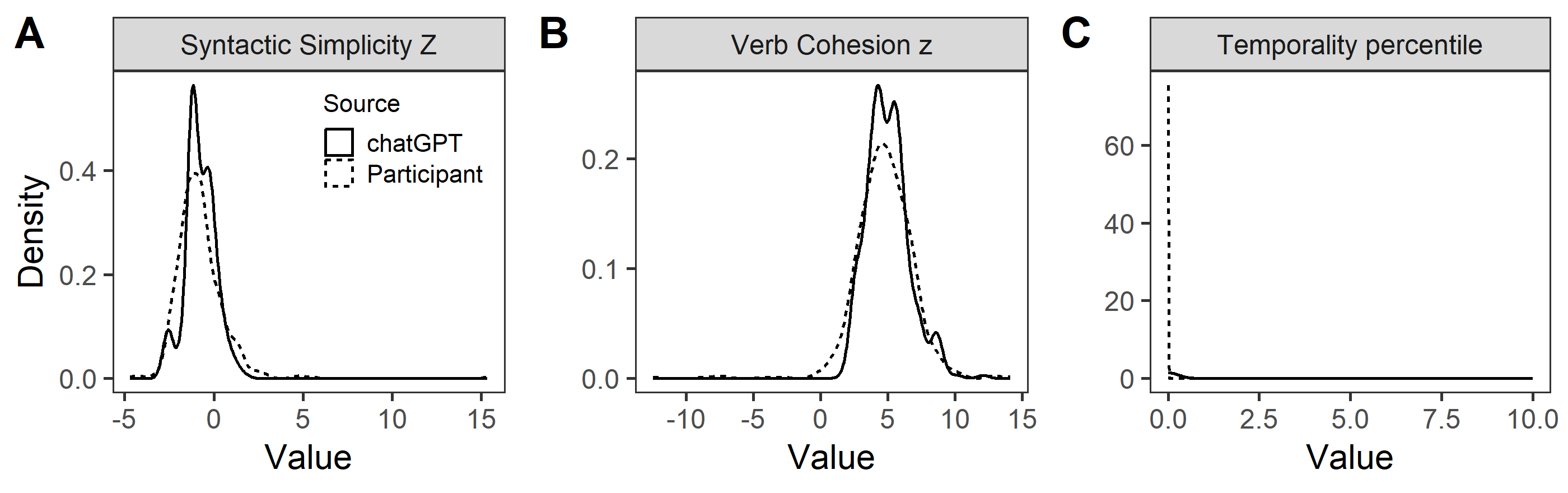}
    \caption{Distribution plots for the Z scores of syntactic simplicity (\textbf{A}), verb cohesion (\textbf{B}), and the percentile score for temporality (\textbf{C}).}
    \label{fig:a2_dists}
\end{figure}

Next, we use the hierarchical clustering approach described above to evaluate which features are most informative for classifying the data. We identified 11 features. These features are word length in syllables, noun phrase density, the type token ratio for content word lemmas, and 8 different text easability metrics. Five of these text easablity metrics were the same as those for analysis 1 (narrativity, word concreteness, referential cohesion, deep cohesion, connectivity, and temporality). The three newly identified features are the z score for syntactic simplicity, the z score for verb cohesion, and the percentile score for temporality. Distribution plots for these variables are illustrated in Figure \ref{fig:a2_dists}.

Results from fitting a ridge classifier with these features are presented in the bottom half of Table ~\ref{tab:analysis2}. Performance on this model is reduced compared to the full model presented in Analysis 1 (Dependent $t$-test: Accuracy: $t = -7.51$, Precision: $t = -7.83$, Recall: $t = -7.60$, $df = 9$, $p < 0.001$). However, this analysis still illustrates reasonably high levels of classification performance. Results from this feature analysis model are comparable to those reported in the feature analysis model in Analysis 1 (dependent $t$-test: Accuracy: $t = 1.35$, Precision: $t = 1.43$, Recall: $t=1.71$, $df = 9$, $p > 0.05$). Classification accuracy does not depend on word count or sentence length.

\subsection{Analysis 3}

Results from analysis 2 illustrated that some remaining descriptive linguistic features may be contributing to high model performance. Notably, word length in syllables was one of the features identified by the feature analysis method. In analysis 3, we conduct a more stringent test and remove all descriptive linguistic features from the data. In addition to total word count, sentence length, and sentence length standard deviation, we remove four features associated with word length.

Results from this model are illustrated in the top of Table~\ref{tab:analysis3}. Results from this model are comparable to those reported in Tables ~\ref{tab:analysis1} ~\ref{tab:analysis2} (Repeated measures ANOVA: Accuracy: $F = 0.12$, Precision: $F = 0.16$, Recall: $F = 0.17$, $df_n = 2, df_d = 18, p > 0.05$). Results indicate high performance at classifying the data as either created by subjects or created by chatGPT.

\begin{table}[h]
\begin{center} 
\caption{Analysis 3 Results} 
\label{tab:analysis3} 
\vskip 0.12in
\begin{tabular}{lll} 
\hline
Model & Metric &  Score: Mean (sd) \\
\hline
Classification & Balanced Accuracy & 0.89 (0.08) \\
Classification & Weighted Precision & 0.90 (0.06) \\
Classification & Weighted Recall & 0.89 (0.05) \\
\hline
Feature (11) & Balanced Accuracy & 0.60 (0.02) \\
Feature (11) & Weighted Precision & 0.67 (0.02) \\
Feature (11) & Weighted Recall & 0.69 (0.02) \\
\hline
Feature (19) & Balanced Accuracy & 0.71 (0.06) \\
Feature (19) & Weighted Precision & 0.76 (0.04) \\
Feature (19) & Weighted Recall & 0.77 (0.03) \\
\hline
\end{tabular} 
\caption*{\emph{Note.} \textbf{Top:} Results from ridge classifier fit on 71 Coh-metrix features (all descriptive linguistic features are removed). \textbf{Middle:} Results from ridge classifier fit on 11 Coh-metrix features selected via a hierarchical clustering approach based on Analysis 3. \textbf{Bottom:} Results from ridge classifier fit on 19 Coh-metrix features selected via a hierarchical clustering approach based on Analysis 3. All result metrics are means and standard deviations from 10-fold cross validation. }
\end{center} 
\end{table}

The 11 features identified using the clustering approach on this model were: noun phrase density, the type token ratio for content word lemmas, and 9 different text easability components. The easability components were the z-scores for narrativity, word concreteness, referential cohesion, deep cohesion, connectivity,  temporality, verb cohesion, syntactic simplicity, and the percentile score for temporality. All of these features were identified in at least one of the feature analyses conducted in analysis 1 and 2. 

Results from training a ridge classifier on these 11 features are illustrated in the middle of Table \ref{tab:analysis3}. Results indicate reduced classification performance compared to the feature analysis model results reported in Table ~\ref{tab:analysis2} (Repeated measures ANOVA: Accuracy: $F = -6.3$, Precision: $F = -5.58$, Recall: $F = -5.02$, $df = 9$, $p < 0.01$). These results illustrate that removing all descriptive linguistic features from the model reduces model performance. 

\begin{figure}[h]
    \includegraphics[width = \columnwidth]{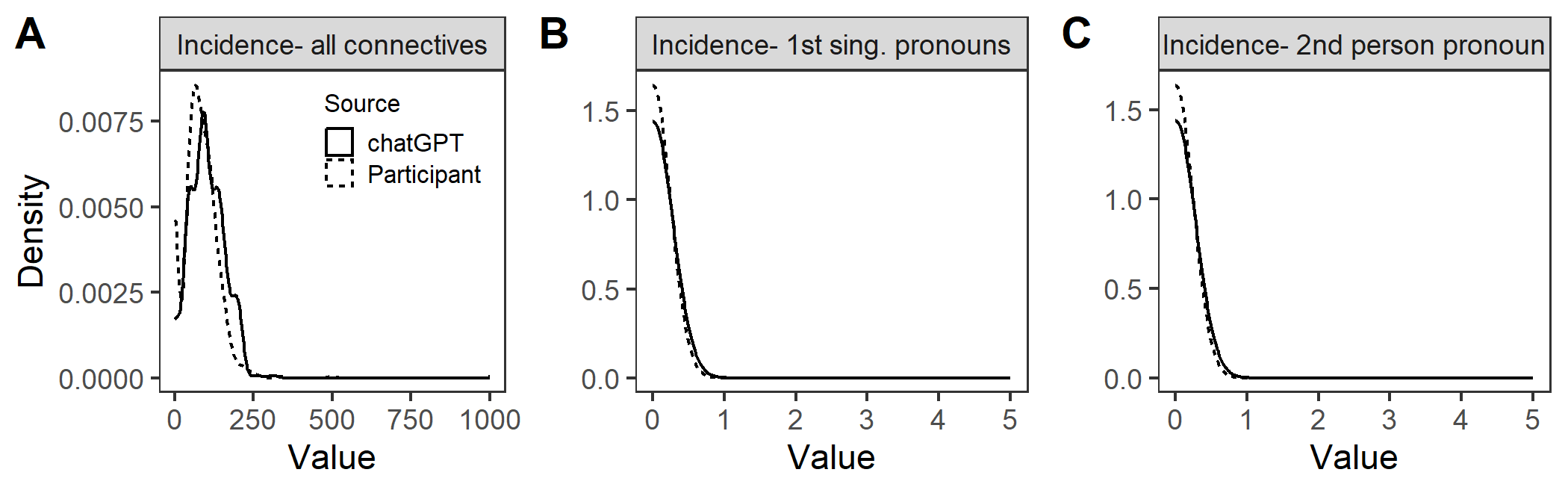}
    \includegraphics[width = \columnwidth]{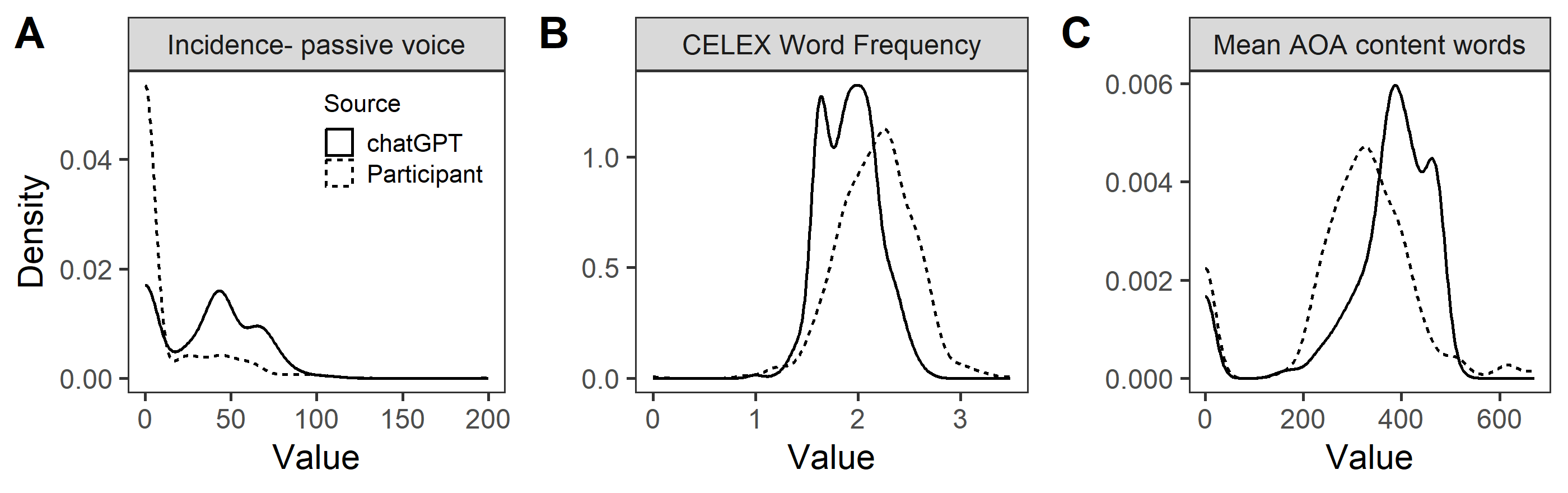}
    \includegraphics[width = 0.66\columnwidth]{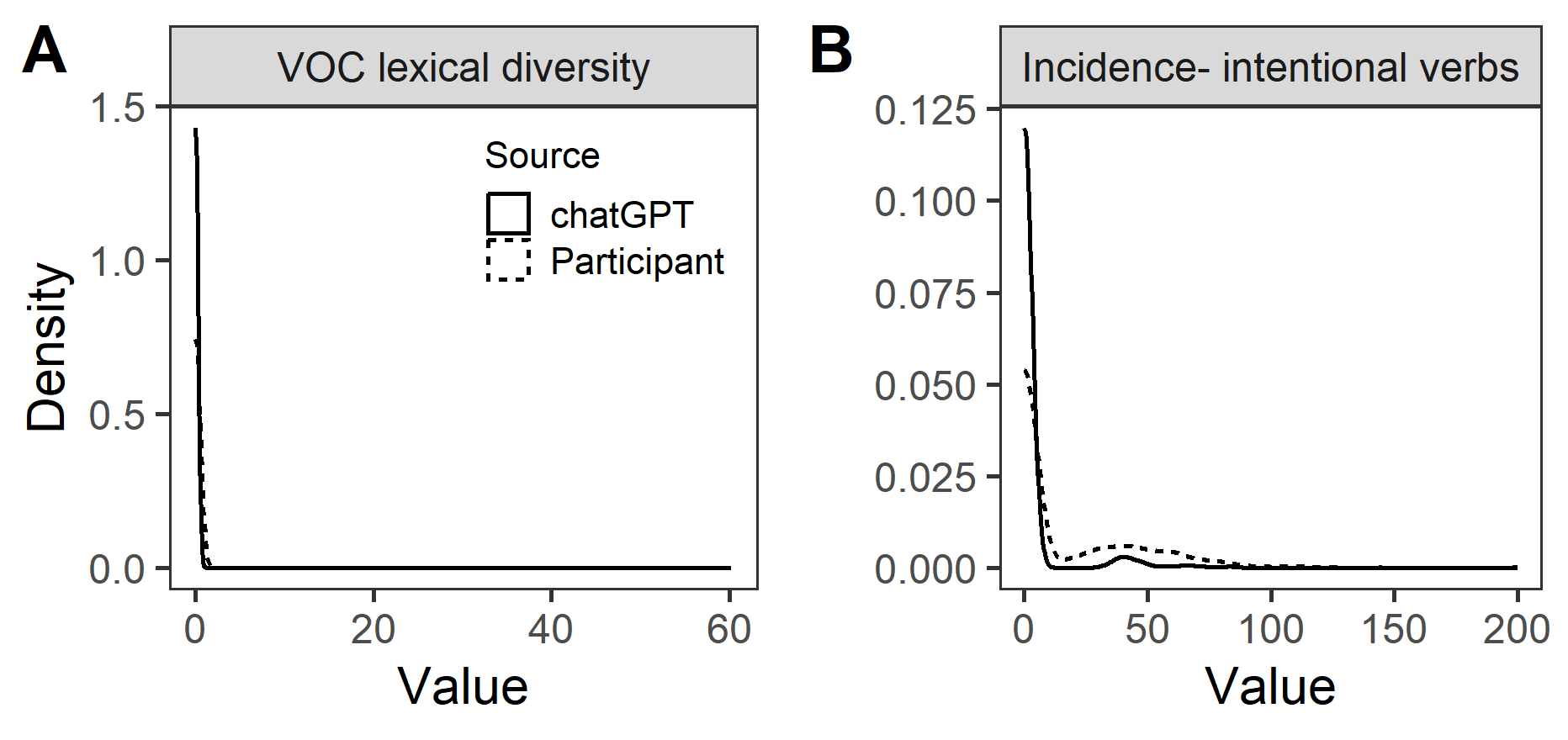}
    \caption{Distribution plots for variables identified in analysis 3. \textbf{Top row:} Incidence of all connectives (\textbf{A}), first person singular pronouns (\textbf{B}), and second person pronouns (\textbf{C}). \textbf{Middle row:} Incidence of passive voice (\textbf{A}), CELEX word frequency (\textbf{B}), and mean AOA for content words (\textbf{C}). \textbf{Bottom row:} VOC lexical diversity (\textbf{A}), incidence for intentional verbs (\textbf{C}). }
    \label{fig:a3_dists}
\end{figure}

We perform a follow-up analysis to estimate how many features are required to generate results that are comparable to those of the feature analyses reported in Tables ~\ref{tab:analysis1} and ~\ref{tab:analysis2}. We execute this by adjusting the cophenatic distance used to select candidate features. Results from this procedure suggest that using 19 features (selected with a cophenatic distance of 0.95) creates a model with similar performance. Results from this model appear in the bottom of Table \ref{tab:analysis3}. In addition to the 11 features identified previously, the 19 features include: incidence rates for all connectives, first person singular pronouns, second person pronouns, and passive voice; the mean word frequency for content words (derived from CELEX \citeNP{baayen_celex_1996}); the mean age of acquisition for content words, lexical diversity as measured by the VOCD \cite{McCarthy2010}; and the incidence rate for intentional verbs. Distribution plots for the 8 new features are illustrated in Figure ~\ref{fig:a3_dists}.

\begin{table}[h]
\begin{flushleft} 
\caption{Summary of Psycholinguistic Differences} 
\label{disc:style} 
\vskip 0.12in
\begin{tabular}{p{1.31in}p{.93in}p{.64in}} 
\hline
Feature & $t$-test ($df$) & Levene's $F$\\
\hline
Narrativity $Z$ & $21.41 (1099)$** & $128.53$**\\

Referential cohesion $Z$ & $8.36 (1084.9)$** & $57.43$**\\

Deep cohesion $Z$ & $5.69 (1059.1)$** & $12.96$** \\

Verb cohesion $Z$ & $-2.31 (960.99)$* & $13.23$** \\

Temporality $Z$ & $4.10 (799.01)$* & $5.98$*\\

Temporality \%  & $2 (734)$* & $2$ \\

Word Concreteness $Z$ & $-.82 (839.87)$ & $3.22$ \\

Explicit connectives $Z$ & $7.19 (787.57)$** & $3.39$ \\

Syntactic simplicity $Z$ & $-.38 (1036.4)$ & $18.17$** \\

Noun phrase density & $2.64 (750.74)$* & $0.19$ \\

TTR content word & $3.19 (1061.7)$* & $5.44$* \\

Connectives I & $-5.11 (832.1)$** & $0.55$ \\

1st singular pronouns I & $4.99 (734)$** & $12.44$** \\

2nd pronouns I & $5.39 (762.86)$**& $14.68$** \\

Passive voice I  & $-13.41 (637.42)$** & $79.02$** \\

CELEX word freq. & $12.99 (955.93)$** & $25.77$** \\ 

AOA content words & $-7.18 (780.73)$** & $4.59$*  \\

VOCD lexical diversity & $1 (734)$ & $.49$ \\

Intentional verbs I & $10.77 (1081.5)$** & $72.72$** \\

\hline

\end{tabular} 
\caption*{\emph{Note.} Inferential statistics for identified features. ** = $p < 0.01$ * = $p < 0.05$, I = incidence score. All Levene's tests have $df = 1099$.}
\end{flushleft} 
\end{table}

\section{General Discussion}
Results from our analyses suggest that long-form analogies generated by chatGPT differ from those generated by human participants in terms of both descriptive linguistic properties and underlying psycholinguistic properties. It is possible that differences in descriptive linguistic features may be unique to our dataset. However, there are several fruitful areas for future study. 

Our results suggest that chatGPT and the human participants tend to select different types of words in responses. Moreover, for many features, participant summaries exhibit more variance in the words they select. Previous work on detecting text generated by LLMs has noted that LLM generated text tends to have distributional or statistical properties that differ from that of text generated by humans \cite{mitchellDetectGPTZeroShotMachineGenerated2023, shen_textdefense_2023, varshney_limits_2020}. We extend this finding by demonstrating that human authors select different function words (illustrated in incidence rates for explicit connectives, all connectives, and pronouns). Moreover, we demonstrate that this variance is reflected in multiple parts of speech for content words (intentional verbs and verb cohesion).

Second, our results suggest differences between the two sources that extend beyond individual word choice. We demonstrate that the two groups differ in style dimensions like syntactic simplicity, passive voice, and narrativity. Differences in narrativity are particularly interesting given that chatGPT is optimized for dialogue, which one might hypothesize would lead to more narrative-like output. Lastly, we demonstrate that the two groups differ in the use of devices that assist readers with developing a complete understanding of the text (deep, referential, and verb cohesion; connectivity, and temporality).  

\subsection{Limitations and Future Research}
Despite our efforts to control the data set, due to differences in the length of the analogies generated by chatGPT and the participants, we have an unbalanced dataset with more sentences from human subjects than chatGPT. Because chatGPT is being actively updated \footnote{See iterative deployment:  https://openai.com/blog/chatgpt/}, it is unclear whether a user is interacting with the same model on a different day. Thus, we opted to not add any additional examples that were generated on a different day that might introduce unwanted systematic variation into the dataset. Second, we did not give chatGPT any specific stylistic instructions. For example, we did not specify the age or educational level of the purported author. Equally relevant, we have not yet conducted domain-specific content analyses that may prove diagnostic.  However, given the broad domain content resources available to chatGPT, we may find that human writers draw on less content.  

Prior research performed sentence level analyses on a graded subset corpus \cite{wijesiriwardene_towards_2022}. Although their results are substantially less dramatic, the conclusions are convergent. Language skills appear to determine manual grades, with broader impact implications for pedagogical assessment practices.  
Finally, we have not yet identified the psychological processes that result in the documented differences.     

\section{Conclusion}
In this work, we examined the psycholinguistic properties of long-form analogies generated by chatGPT. Advances in natural language processing and deep learning have led to the development of large language models that can generate convincing and fluent text. This apparent fluency prompts the need for evaluation techniques that can capture subtle, underlying features of skilled language use. Psycholinguistic features are well-suited to this task and have the added benefit of explainability. We demonstrate that features drawn from established psycholinguistic research can identify the differences between human and computationally generated text in a complex reasoning exercise.

\section{Acknowledgments}
We thank the anonymous reviewers for their comments and suggestions. We have no funding to disclose for this project. The first author has no relation to EngageFast learning. The second author has previously collaborated with the founder of EngageFast learning.

The first author is also affiliated with the Air Force Research Laboratory and Oak Ridge Institute for Science and Education. This research is a personal project. The views expressed are those of the authors and do not necessarily reflect the official policy or position of the Department of the Air Force, the Department of Defense, or the U.S. government. Cleared AFRL-2023-2262.

\bibliographystyle{apacite}
\setlength{\bibleftmargin}{.125in}
\setlength{\bibindent}{-\bibleftmargin}
\bibliography{references}
\end{document}